\documentclass[conference]{IEEEtran}
\usepackage{graphicx}
\usepackage{amsmath}
\usepackage{booktabs}
\IEEEoverridecommandlockouts
\ifCLASSINFOpdf
\else
\fi

\begin{document}
\title{Exploring Differences in Interpretation of Words Essential in Medical Expert-Patient Communication}
% * <bulerias.snow.toy@gmail.com> 2016-03-23T13:15:35.708Z:
%
% Modified as suggested by 1st reviewer
%
% ^ <bulerias.snow.toy@gmail.com> 2016-04-04T18:10:16.995Z.
%\title{Exploring Differences in Interpretation of Words Essential in Medical Treatment by Patients and Medical Professionals}
\author{\IEEEauthorblockN{J. Navarro$ ^{a} $, C. Wagner$ ^{a} $, U. Aickelin$ ^{b} $, L. Green$ ^{c} $ and R. Ashford$ ^{c} $\\}
\IEEEauthorblockA{$ ^{a} $Lab for Uncertainty in Data and Decision Making (LUCID) and Horizon,\\School of Computer Science, University of Nottingham, UK\\$ ^{b} $The University of Nottingham Ningbo\\
$ ^{c} $Nottingham University Hospitals, UK\\Email: \{psxfjn,christian.wagner,uwe.aickelin\}@nottingham.ac.uk\\Email: lynsey.green@nuh.nhs.uk, rashford@nhs.net}\thanks{This work was partially funded by the RCUK grant EP/M02315X/1 From Human Data to Personal Experience.}}
%\and
%\IEEEauthorblockN{R. Ashford and L. Green}
%\IEEEauthorblockA{rashford@nhs.net\\lynsey.green@nuh.nhs.uk}

%\thanks{J. Navarro, C. Wagner and U. Aickelin are with the School of Computer Science, University of Nottingham, UK. Email: \{psxfjn,christian.wagner,uwe.aickelin\}@nottingham.ac.uk}
%\thanks{C. Wagner is also with the Lab for Uncertainty in Data and Decision Making (LUCID) and Horizon, School of Computer Science}
%\thanks{R. Ashford and L. Green are with the Nottingham University Hospitals, UK. Email: rashford@nhs.net, lynsey.green@nuh.nhs.uk}}
\maketitle

\begin{abstract} 
In the context of cancer treatment and surgery, quality of life assessment is a crucial part of determining treatment success and viability. In order to assess it, patient-completed questionnaires which employ words to capture aspects of patients’ well-being are the norm. As the results of these questionnaires are often used to assess patient progress and to determine future treatment options, it is important to establish that the words used are interpreted in the same way by both patients and medical professionals. In this paper, we capture and model patients’ perceptions and associated uncertainty about the words used to describe the level of their physical function used in the highly common (in Sarcoma Services) Toronto Extremity Salvage Score (TESS) questionnaire. The paper provides detail about the interval-valued data capture as well as the subsequent modelling of the data using fuzzy sets. Based on an initial sample of participants, we use Jaccard similarity on the resulting words models to show that there may be considerable differences in the interpretation of commonly used questionnaire terms, thus presenting a very real risk of miscommunication between patients and medical professionals as well as within the group of medical professionals.
\end{abstract}


\IEEEpeerreviewmaketitle

\section{Introduction}
Fuzzy Set Theory (introduced in \cite{zadeh1965fuzzy}) has provided a framework for modelling of uncertainties in a vast field of applications through Fuzzy Sets (FSs). In particular, one of the most recent fields of study is a paradigm called Computing with Words (CW) \cite{Zadeh1996}, which has established a methodology where words are used in place of numbers for computing and reasoning.

As part of this recent interest, modelling the subjective meaning of words and perceptions of people has been investigated and a number of techniques have been proposed, such as the Interval Approach (IA) \cite{Feilong2008a}, the Enhanced Interval Approach (EIA) \cite{Coupland2010,Wu2012a}, and the Interval Agreement Approach (IAA) \cite{Wagner2014}. Such techniques allow the creation of FS models (from data) for words and/or concepts in order to subsequently perform computation and reasoning. The potential applications are diverse such as: recommendation systems (e.g., \cite{McCulloch2015}) and decision support systems (refer to \cite{Herrera2009} for an overview). In this study, we use IAA since it avoids making assumptions about the distributions of the models generated.% While IA and EIA approaches are valuable options for modelling of survey-based data, they require making assumptions about the data by depending on specific FS models.
%While IA and EIA approaches are able of representing words through Interval Type-2 FSs (IT2 FSs) assuming a shape (e.g., trapezoidal or shoulder) of the Footprint Of Uncertainty (FOU), they require data pre-processing (e.g., outlier removal) and model all types of uncertainty in the FOU of IT2 FSs. In contrast, the IAA avoids assumptions about the shapes of the FS membership functions and differentiates between different types of uncertainty (e.g., intra-participant, inter-participant), specifically intra and inter source uncertainty.

In medicine, the capture of information from patients' and medical professionals' perceptions of easiness/difficulty to perform daily activities commonly involves uncertainty due to a number of reasons: variability in people's perceptions throughout the day, mood changes, experience, training, disagreement between groups of people, etc. Several studies have shown that the use of representations of medical data based on FSs (e.g., fuzzy Apgar score \cite{Shono1992}, Cadiag-2 \cite{Ciabattoni2009}) is valuable when uncertainty is present.

In this paper, we present an initial study focusing on two aims: on the one hand, we analyse if the words commonly used in this specific area of medicine, specifically the linguistic descriptors used in the Toronto Extremity Salvage Score (TESS) have the same meaning to different groups of people involved (e.g., medical professionals and patients) by comparing the FS models generated with the Jaccard similarity measure \cite{Jaccard1908}. On the other hand, we explore the practical application of the IAA \cite{Wagner2014,Miller2012,Wagner2013b} to capture interval-valued data and generate FS models associated with such words.

The paper is structured as follows: Section II provides background on TESS, the importance of having standard words in communication between doctors and patients, the IAA method and the Jaccard similarity measure. In Section III, we provide detail about the interval-valued data captured from 3 different groups of people surveyed (surgeons, physiotherapists and patients) as well as a demonstration of their processing through IAA. Models generated from the data, as well as comparisons between different sources are presented in Section IV. Finally, Section V provides a discussion of the results obtained and their implications both in technical and in application terms while Section VI presents the conclusions and challenges/directions for future work.

\section{Background}
This section presents the TESS questionnaire and discusses the importance of words in medicine. Later, we introduce the IAA used to generate FSs from interval-valued responses as well as the similarity measure used to compare the resulting sets.

 \subsection{Toronto Extremity Salvage Score (TESS)}
\begin{figure}[h]
\centering
\includegraphics[scale=0.4]{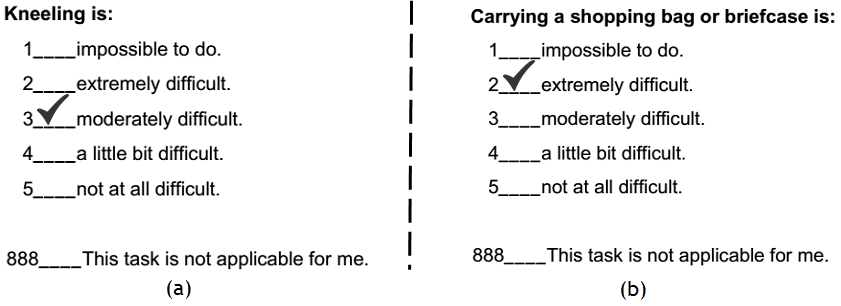}
\caption{Two sample TESS items: (a) item taken from the lower extremity questionnaire, (b) item taken from the upper extremity questionnaire.}
\label{fig:tess}
\end{figure}

The Toronto Extremity Salvage Score (TESS) is a standard patient-completed questionnaire for assessment of function following sarcoma surgery \cite{Davis1996}. It was developed to monitor the effects of therapeutic interventions on patients undergoing sarcoma surgery on extremities. TESS is commonly administered at four time points: the first session (which is commonly before surgery) and 12, 18 and 24 months from then on. The TESS consists of a 30-item lower limb and 29-item upper limb questionnaire that allows participants to pick a selection of responses representing their perceptions about the extent of difficulty to perform daily activities based on a Likert scale. In Fig. \ref{fig:tess}, we show a fragment of the TESS with two items taken from the lower and upper extremity questionnaires respectively.

Note that the questionnaires use the same linguistic descriptors e.g.: \textit{Impossible to do} and \textit{Extremely difficult}. Commonly, after having being completed by the patient, the whole set of answers is used to generate a standardized score ranging from 0 to 100. This evaluated TESS is finally analysed by surgeons/physiotherapists in order to measure changes in physical functions over time as well as evaluating the need for treatment intervention, assistive devices, job modifications, etc.

\subsection{Importance of Words in Medicine}%use of terms in the medical field (Rob and Lynsey could help here). % TESS, is composed up of a numerical ordinal scale with a descriptive level of function.
While words in questionnaires are convenient, the challenge of dealing with different interpretations of words (``words mean different things to different people'') has previously been identified in Medicine in the context of communication with patients (e.g., patient consent, risk communication) \cite{Merz1991}. Therefore, there is a motivation for reducing miscommunication since growing evidence indicates that errors in communication can give rise to an increase of risk in terms of clinical morbidity and mortality \cite{Coiera}. In the context of consent specifically, effective communication is the basis for informed patient consent for medical treatment. More specifically, doctors must empathise with the emotions of the patient and also, explain the possible outcomes and associated risks\cite{Paling2003a}. %Nevertheless, recent findings on the perception of risks and benefits from a psychological perspective have suggested that patients just extract the gist of any information -not the detail- to make decisions \cite{Lloyd2001}.

Moreover, also in the context of consent, the European Union suggests to use a standardised vocabulary (``very common'', ``common'', ``uncommon'', ``rare'', and ``very rare''). However, patients' interpretations of these terms do not seem to correlate with the probabilities that they were intended to convey \cite{Berry2003} since abundant evidence points out that descriptive terms reflect the speaker's perspective, with the patient often understanding the risks to be of a totally different order of magnitude \cite{Merz1991}. In addition, different countries probably bring different shades of meaning to various descriptions \cite{Berry2003}. For these reasons, Paling \cite{Paling2003a} recommends to discuss with colleagues (at a local and national level) the use of a standardised vocabulary of descriptive words so that miscommunication is reduced.

%http://www.ncbi.nlm.nih.gov/pmc/articles/PMC200818/
%http://www.ncbi.nlm.nih.gov/pmc/articles/PMC61430/
While both the importance and the challenge around miscommunication of/through words has been explored in detail in the context of patient consent, the same is not the case in the context of patient treatment and quality of life assessment (i.e., as in the TESS). Thus, this paper employs recently developed data capture and modelling techniques to explore  and numerically assess potential variations in meaning of key words by medical professionals and patients.

\subsection{Interval Agreement Approach}
In \cite{Wagner2014}, the Interval Agreement Approach (IAA) is introduced as a method for generating FSs from interval-valued data representing uncertainty in people's opinions/perceptions. It is built on top of the work presented in \cite{Miller2012}, where an agreement-based method \cite{Wagner2011} of capturing interval-valued survey data is demonstrated. Also, in \cite{Wagner2013b} its practical application along with the use of a similarity measures to relate attribute word models to concept models is explored.

The IAA considers two types of intervals in the process of capturing responses: crisp (no uncertainty about the interval endpoints) and uncertain (each endpoint modelled itself as a crisp interval). Also, it considers two types of uncertainty to be modelled through different dimensions of the resultant FSs, namely inter-source (variation among a group of participants) and intra-source (variation in the opinion of a particular participant). Depending on the data, the IAA can generate:
\begin{itemize}
	\item Type-1 FSs (T1 FSs). In this case, crisp intervals and either inter- or intra-source uncertainty are modelled in the primary degree of membership $ y $ (or $ \mu $) by combining multiple intervals,
	\item Interval type-2 (IT2 FSs). In this case, uncertain intervals and also, either inter- or intra-source uncertainty is modelled in the primary degree of membership $ y $ (or $ \mu $) by combining multiple intervals,
	\item General type-2 FS based on zSlices \cite{Wagner2010a}. In this case, both inter- and intra-source uncertainty are being modelled through the primary $ y $ (or $ u $) and secondary $ z $ (or $ \mu $) degrees of membership.
\end{itemize}
 In this paper, we conduct a single iteration of a survey with multiple participants. Thus, we are focusing on capturing inter-participant uncertainty through crisp intervals and will limit the further description of IAA to the case of T1 FS generation.

\begin{figure}[h]
	\centering
	\includegraphics[scale=0.45]{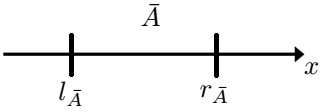}
	\caption{Crisp interval $ \bar { A }  $.}
	\label{fig:crispInterval}
\end{figure}

\begin{figure}[h]
	\centering
	\includegraphics[scale=0.35]{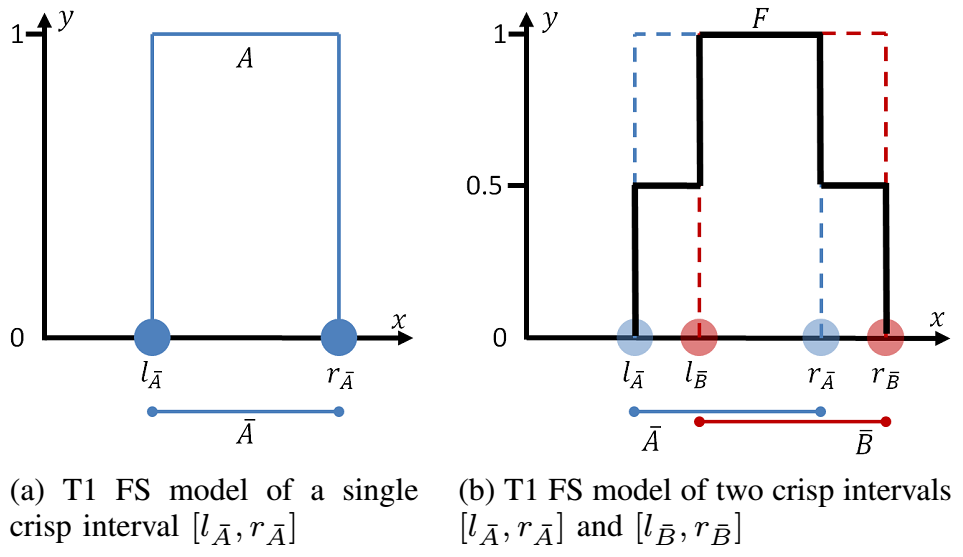}
	\caption{Generation of T1 FSs from crisp interval(s) using the IAA. Figure taken from \cite{Wagner2014}.}
	\label{fig:sampleT1FSs}
\end{figure}

Let $ \bar { A }  $ be a crisp interval with the left and right endpoints $ { l }_{ \bar { A }  } $ and $ { r }_{ \bar { A }  } $ (see Fig. \ref{fig:crispInterval}a). For a given set of sources (e.g., a group of patients), a T1 FS is created on the basis of the provided crisp interval(s) (see Fig. \ref{fig:sampleT1FSs}) representing the agreement between different participants' opinions/responses. The degree of membership $ y $ of the set over the survey domain $ x $ captures the number of intervals overlapping at a particular point. Figures \ref{fig:crispInterval}a and \ref{fig:crispInterval}b show the case of generating a T1 FS for single and multiple intervals respectively using the IAA.

Given a series of intervals $ { \bar { A }  }_{ n } $ to be combined in a T1 FS $ A $, $ n\in \left\{ 1,...,N \right\} $, where $ N $ is the number of intervals, the membership function of $ A $ (denoted by $ \mu(A) $) is described as shown in (\ref{eq:IAA}).

\begin{equation}\label{eq:IAA}
\begin{aligned}
	{ \mu  }_{ A }\left( x' \right) =\frac { \left( \sum _{ i=1 }^{ N }{ { \mu  }_{ \bar { { A }_{ i } }  }\left( x' \right)  }  \right)  }{ N } ,
	\end{aligned}
\end{equation}

\begin{center}
	where: $ { \mu  }_{ \bar { { A }_{ i } }  }\left( x'  \right) =\begin{cases} 1\qquad { l }_{ \bar { { A }_{ i } }  }\le x' \le { r }_{ \bar { { A }_{ i } }  } \\ 0\qquad else \end{cases}. $
\end{center}

\subsection{Similarity Measures}
Similarity measures are functions used in (fuzzy) set theory to compare crisp and FSs. This section provides a brief overview of these methods focusing on the measure selected for this article.

A similarity measure $ s : S(A,B) \rightarrow [0,1]  $ is a function which assigns a similarity value $ s $ to a pair of fuzzy sets $ (A,B) $ that indicates the degree to which the FSs $ A $ and $ B $ are similar \cite{Setnes1998}. One of the most used methods to measure similarity in both crisp and fuzzy set theory is the Jaccard similarity coefficient \cite{Jaccard1908} which, in the case of type-1 fuzzy sets, can be expressed as:
\begin{equation}\label{eq:jaccard}
{ S }_{ J }^{ FS }\left( A,B \right) =\frac { \sum _{ i=1 }^{ N }{ \min { \left( { \mu  }_{ A }\left( { x }_{ i } \right) ,{ \mu  }_{ B }\left( { x }_{ i } \right)  \right)  }  }  }{ \sum _{ i=1 }^{ N }{ \max { \left( { \mu  }_{ A }\left( { x }_{ i } \right) ,{ \mu  }_{ B }\left( { x }_{ i } \right)  \right)  }  }  },
\end{equation}
%\ref{eq:jaccard}
where $ N $ is the total number of discretisations along the \textit{x}-axis, and $  x \in X $ is the domain of the membership functions associated with the FSs $ A $ and $ B $. The result indicates how similar $ A $ is to $ B $, with 1 indicating that both FSs are identical and 0 that they are disjoint.

% Note that IEEE typically puts floats only at the top, even when this
% results in a large percentage of a column being occupied by floats.

% An example of a double column floating figure using two subfigures.
% (The subfig.sty package must be loaded for this to work.)
% The subfigure \label commands are set within each subfloat command, the
% \label for the overall figure must come after \caption.
% \hfil must be used as a separator to get equal spacing.
% The subfigure.sty package works much the same way, except \subfigure is
% used instead of \subfloat.
%
%\begin{figure*}[!t]
%\centerline{\subfloat[Case I]\includegraphics[width=2.5in]{subfigcase1}%
%\hfil
%\subfloat[Case II]{\includegraphics[width=2.5in]{subfigcase2}%
%\label{fig_second_case}}}
%\caption{Simulation results}
%\label{fig_sim}
%\end{figure*}
%
% Note that often IEEE papers with subfigures do not employ subfigure
% captions (using the optional argument to \subfloat), but instead will
% reference/describe all of them (a), (b), etc., within the main caption.

% An example of a floating table. Note that, for IEEE style tables, the 
% \caption command should come BEFORE the table. Table text will default to
% \footnotesize as IEEE normally uses this smaller font for tables.
% The \label must come after \caption as always.
%

\section{Comparing the Perception of Linguistically Expressed Functional Outcomes}
In this section we describe the data collection conducted and provide an example of processing the interval-valued responses to generate FS models for later analysis in Section \ref{section4}. 
\subsection{Overview}
We surveyed thirty-six participants (12 sarcoma surgeons, 12 physiotherapists and 12 patients undergoing lower limb salvage surgery) on the 5 levels of each score from the TESS items: ``\textit{impossible to do}'',``\textit{extremely difficult}'', ``\textit{moderately difficult}'', ``\textit{a little bit difficult}'', and ``\textit{not at all difficult}''. The sarcoma surgeons completed the questionnaires at the British Orthopaedic Association Annual Scientific Meeting and the physiotherapists at the Sarcoma Physiotherapist Networking Event, whereas the patients were given instructions during their respective medical consultations and completed the questionnaires prior to leaving the hospital following their consultation.%The medical professionals completed the questionnaires in the Nottingham City Hospital facilities whereas the patients were given instructions during their respective medical consultations and were free to complete the questionnaires at home.

While the questionnaires were provided to all the participants within the same time frame, in the specific case of patients is noteworthy mentioning that their medical consultations were conducted at different points with regard to their surgery, i.e., some of them were completed before surgery intervention and some others after surgery (up to 18 months). Note that, as the questionnaire focused on the interpretation of the linguistic terms, and not on the patients' actual condition, the timing is not considered relevant.
%Explain the problem
%In general terms, intra-participant uncertainty (either from patient or medical professional) is present in Likert-based questionnaires (such as TESS) for a number of reasons: variation amongst the individual opinions of groups of participants, mood changes, different backgrounds, etc. Also, due to the nature of such questionnaires, the value assigned to a Likert item has no objective numerical basis so their perceptions might be different.
%To our initial perspective, the opinions about the linguistic descriptors related to the easiness/difficulty to perform daily tasks involved in TESS, disagree considerably amongst professionals and patients. We attribute this to two reasons: (1) The words mean different things to different people and some of them imply describing with a more vague/precise extent than others; (2) The professional background can play an important role when interpreting/describing extremity functions given that the expectations about the therapy effects might vary due to different criteria, years of experience, etc.

\subsection{Processing the real responses}
As an initial part of this study, we developed a questionnaire with a continuous interval-valued scale for each item. We provided instructions to the participants to draw an ellipse around the appropriate extent of each of the 5 linguistic descriptors (words) used in the TESS score. The position and width of the ellipse on the scale indicated the difficulty level and the extent of the uncertainty perceived by the participant.

After collating the questionnaires, we proceeded to extract the interval-valued data and to model the inter-participant agreement from the different groups: patients, surgeons, physiotherapists and a combined model from both surgeons and physiotherapists representing the body of medical professionals. The rationale of doing so, was to analyse the variation in interpretation of TESS items amongst patients and medical professionals. This analysis is presented in Section \ref{section4}.

\subsection{Data Modelling Example}
\begin{figure}[h]
	\centering
	\includegraphics[scale=0.37]{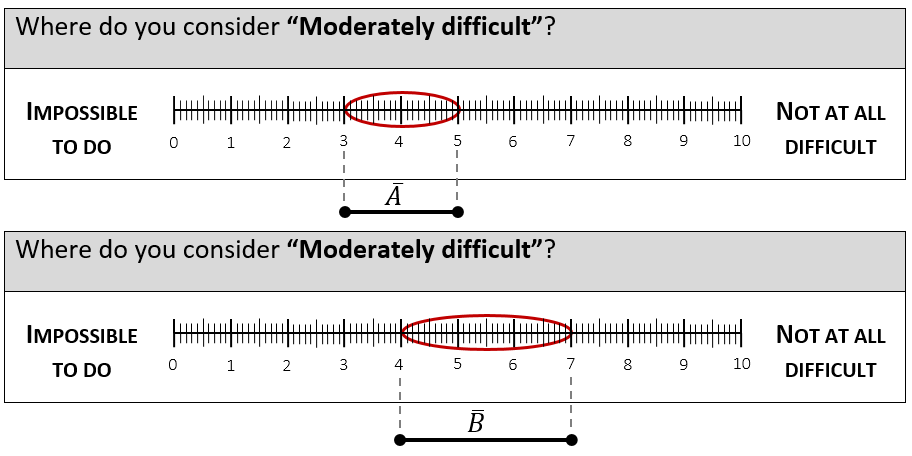}
	\caption{Two sample ellipses from two sources. Each ellipse results in one interval.}
	\label{fig:tessSyntheticExample}
\end{figure}

%generation of the models
For a given subject/concept, consider two intervals $ \bar{A} $ and $ \bar{B} $ generated from the ellipses displayed in Fig. \ref{fig:tessSyntheticExample} where we are showing two possible responses from two different sources/participants ($ N=2 $). Note that the wider the ellipse/interval, the higher the uncertainty in the answer. In other words, this difference in the width indicates how a participant might answer a question s/he is $ a $) relatively certain or $ b $) fairly uncertain of.

From the intervals $ \bar { A } =\left[ { l }_{ \bar { A }  },{ r }_{ \bar { A }  } \right] =\left[ 3,5 \right]   $ and $ \bar { B } =\left[ { l }_{ \bar { B }  },{ r }_{ \bar { B }  } \right] =\left[ 4,7 \right]  $ the FS $ C $ is generated using a discrete representation (see (\ref{eq:fsC})). For space considerations, we only present the calculations for two discrete values $ x = 3$ and $ x = 4$ (see (\ref{eq:x3x4})).

\begin{equation}\label{eq:x3x4}
\begin{aligned}
{ \mu  }_{ C }\left( x=3 \right) =\frac { \left( 1+0 \right)  }{ 2 } =0.5\\ { \mu  }_{ C }\left( x=4 \right) =\frac { \left( 1+1 \right)  }{ 2 } =1
\end{aligned}
\end{equation}

\begin{equation}\label{eq:fsC}
C=0.5/3 + 1/4 + 1/5 + 0.5/6 + 0.5/7
\end{equation}

\begin{figure}[h]
	\centering
	\includegraphics[scale=0.38]{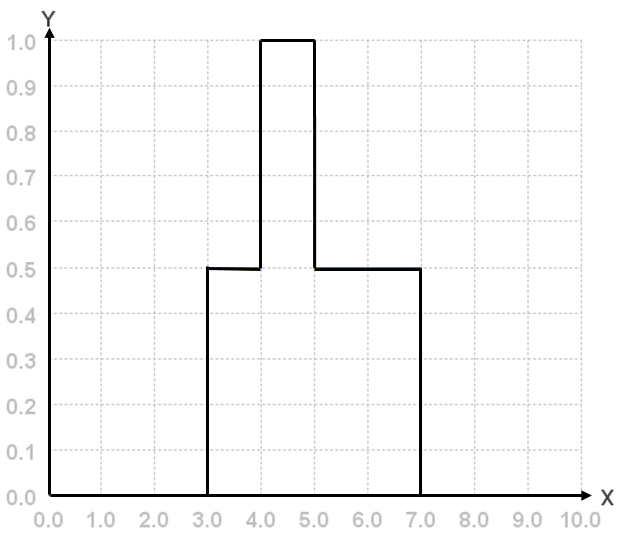}
	\caption{FS model of two intervals $ \left[ { l }_{ \bar { A }  },{ r }_{ \bar { A }  } \right] $ and $ \left[ { l }_{ \bar { B }  },{ r }_{ \bar { B }  } \right] $ from different sources.}
	\label{fig:syntheticFS}
\end{figure}

Finally, the resulting FS $ C $ is depicted in Fig. \ref{fig:syntheticFS}. Note that a non-parametric FS model has been generated and the agreement between the two responses has been modelled in the primary degree of membership $ y $ (commonly called $ \mu $).

\section{Results} \label{section4}
In this section, we present the analysis of the different FSs created with the IAA capturing the agreement between four groups involved in the TESS application, namely: Patients, Physiotherapists, Surgeons and a fourth one created from the combined responses from both Physiotherapists and Surgeons (PS) which together represent the body of ``medical professionals'' interacting with the patients. The rationale of modelling both groups is analysing the extent of agreement/disagreement within the overall group of medical professionals and to provide an overall comparison with the patients.

\subsection{Fuzzy Sets generated}
\begin{figure}[htpb]
	\centering
	\includegraphics[scale=0.43]{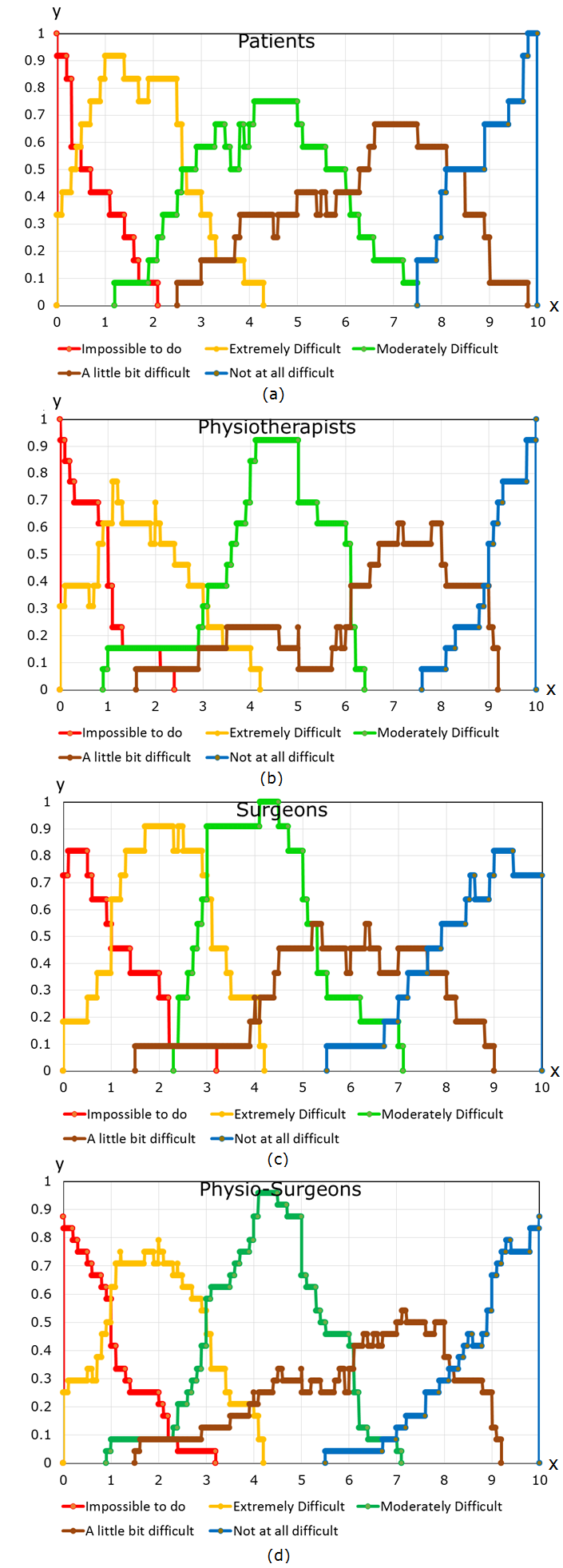}
	\caption{FSs modelling the word concepts: \textit{Impossible to do}, \textit{Extremely difficult}, \textit{Moderately difficult}, \textit{A little bit difficult}, and \textit{Not at all difficult}, generated from different sources: (a) is for patients. (b) is for Physiotherapists. (c) is for Surgeons. (d) is for the combined responses from both Physiotherapists and Surgeons.}
	\label{fig:wordsConceptsFS}
\end{figure}

Figure \ref{fig:wordsConceptsFS}a depicts the FSs for the inter-patient agreement for the 5 linguistic descriptions (words) present in the TESS. As can be seen, the generated FSs have intervals with perfect agreement ($ y=1 $) for the descriptors \textit{Impossible to do} and \textit{Not at all difficult}, whereas for the remaining descriptors \textit{Extremely difficult}, \textit{Moderately difficult} and \textit{A little bit difficult}, there are wider and shorter FSs as a result of lower levels of agreement (overlap) and higher levels of uncertainty.

Figure \ref{fig:wordsConceptsFS}b shows the FS word models generated from the Physiotherapists' responses. It is noteworthy that similarly to the patients' case, these FSs have parts with high agreement such as the descriptors \textit{Impossible to do}, \textit{Moderately difficult}, and \textit{Not at all difficult}. For the remaining descriptors \textit{Extremely difficult} and \textit{A little bit difficult}, it can be seen that there is more disagreement about the extent of such words on the scale. Interestingly, the model for \textit{A little bit difficult} is quasi bi-modal, indicating two different interpretations of the term.

Figure \ref{fig:wordsConceptsFS}c shows the FSs representing the word models generated from Surgeons' responses. For this group, the linguistic descriptor \textit{Moderately difficult} presents the highest level of agreement overall whereas the linguistic descriptors for the extreme cases \textit{Impossible to do} and \textit{Not at all difficult} show wider FSs and lower agreement, representing higher uncertainty perceived by the Surgeons group when using those descriptions in contrast to the previous 2 groups.

In order to provide a comparison from the point of view of Patients vs Medical Professionals, we have created a set of models representing the Medical Professionals by using the responses of both Surgeons and Physiotherapists. The model is shown in Figure \ref{fig:wordsConceptsFS}d. The model is more fine-grained as a result of having more intervals. Low agreement and high uncertainty still relates to the descriptor \textit{A little bit difficult} as in the both original cases and \textit{Moderately difficult} is still the descriptor with most agreement overall.

Overall, it is noteworthy that the model for \textit{A little bit difficult} is the widest and lowest (least agreement) throughout, indicating the term is the least clear in the TESS vocabulary.

\subsection{Comparisons between different groups of people}
While the modelling of interval valued data with the IAA provides framework for visual interpretation of agreement between different groups of people, it is their analysis with similarity measures that provides a numerical one-to-one comparison. Table \ref{tab:similarities} contains all of the comparisons for the different linguistic descriptors based on the Jaccard Similarity measure (see (\ref{eq:jaccard})).

\begin{table}[htbp]
	\centering
	\caption{Similarities between different groups for all words}
	\begin{tabular}{r|rrrr}
		\toprule
		\multicolumn{5}{c}{\textbf{Impossible to do}} \\
		\midrule
		& \textbf{Patient} & \textbf{Physio} & \textbf{Surgeon} & \textbf{PS} \\
		\textbf{Patient} & -     & 0.752 & 0.670 & 0.735 \\
		\textbf{Physio} & 0.752 & -     & 0.710 & 0.838 \\
		\textbf{Surgeon} & 0.670 & 0.710 & -     & 0.847 \\
		\textbf{PS} & 0.735 & 0.838 & 0.847 & - \\
	\midrule
		\multicolumn{5}{c}{\textbf{Extremely difficult}} \\
		\midrule
		& \textbf{Patient} & \textbf{Physio} & \textbf{Surgeon} & \textbf{PS} \\
		\textbf{Patient} & -     & 0.730 & 0.703 & 0.745 \\
		\textbf{Physio} & 0.730 & -     & 0.640 & 0.798 \\
		\textbf{Surgeon} & 0.703 & 0.640 & -     & 0.801 \\
		\textbf{PS} & 0.745 & 0.798 & 0.801 & - \\
\midrule
		\multicolumn{5}{c}{\textbf{Moderately difficult}} \\
		\midrule
		& \textbf{Patient} & \textbf{Physio} & \textbf{Surgeon} & \textbf{PS} \\
		\textbf{Patient} & -     & 0.678 & 0.696 & 0.765 \\
		\textbf{Physio} & 0.678 & -     & 0.614 & 0.785 \\
		\textbf{Surgeon} & 0.696 & 0.614 & -     & 0.786 \\
		\textbf{PS} & 0.765 & 0.785 & 0.786 & - \\
\midrule
		\multicolumn{5}{c}{\textbf{A little bit difficult}} \\
		\midrule
		& \textbf{Patient} & \textbf{Physio} & \textbf{Surgeon} & \textbf{PS} \\
		\textbf{Patient} & -     & 0.683 & 0.707 & 0.726 \\
		\textbf{Physio} & 0.683 & -     & 0.622 & 0.770 \\
		\textbf{Surgeon} & 0.707 & 0.622 & -     & 0.805 \\
		\textbf{PS} & 0.726 & 0.770 & 0.805 & - \\
\midrule
		\multicolumn{5}{c}{\textbf{Not at all difficult}} \\
		\midrule
		& \textbf{Patient} & \textbf{Physio} & \textbf{Surgeon} & \textbf{PS} \\
		\textbf{Patient} & -     & 0.704 & 0.694 & 0.782 \\
		\textbf{Physio} & 0.704 & -     & 0.512 & 0.675 \\
		\textbf{Surgeon} & 0.694 & 0.512 & -     & 0.760 \\
		\textbf{PS} & 0.782 & 0.675 & 0.760 & - \\
		\bottomrule
	\end{tabular}%
	\label{tab:similarities}%
\end{table}%

\begin{figure}[h]
	\centering
	\includegraphics[scale=0.65]{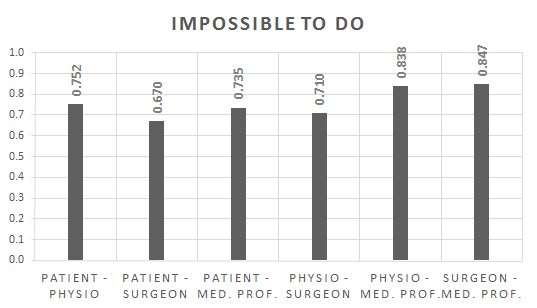}
	\caption{FS similarities between different groups of people for the linguistic descriptor \textit{Impossible to do}.}
	\label{fig:impossible}
\end{figure}

\begin{figure}[h]
	\centering
	\includegraphics[scale=0.65]{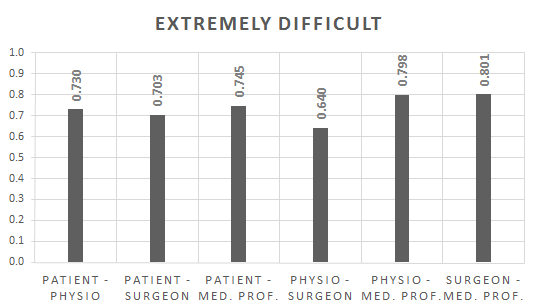}
	\caption{FS similarities between different groups of people for the linguistic descriptor \textit{Extremely difficult}.}
	\label{fig:extremelyDif}
\end{figure}

\begin{figure}[h]
	\centering
	\includegraphics[scale=0.65]{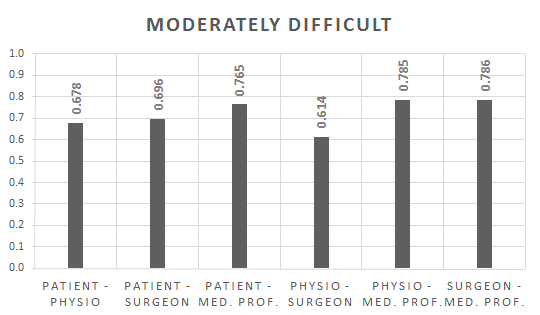}
	\caption{FS similarities between different groups of people for the linguistic descriptor \textit{Moderately difficult}.}
	\label{fig:ModeratelyDif}
\end{figure}

\begin{figure}[h]
	\centering
	\includegraphics[scale=0.64]{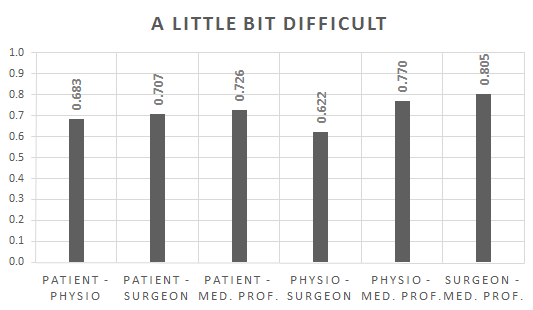}
	\caption{FS similarities between different groups of people for the linguistic descriptor \textit{A little bit difficult}.}
	\label{fig:aLittleBitDif}
\end{figure}

\begin{figure}[h]
	\centering
	\includegraphics[scale=0.64]{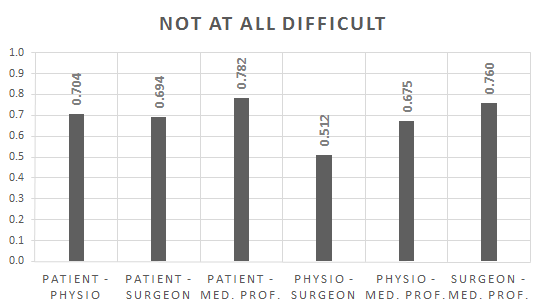}
	\caption{FS similarities between different groups of people for the linguistic descriptor \textit{Not at all difficult}.}
	\label{fig:notAtAllDif}
\end{figure}

Summarizing Table \ref{tab:similarities}, we highlight the least similar FSs:
\begin{itemize}
	\item For the linguistic descriptor \textit{Impossible to do} (see also Fig. \ref{fig:impossible}), the comparison Patients-Surgeons shows the least similar (0.670) models.
	\item For the linguistic descriptor \textit{Extremely difficult} (see also Fig. \ref{fig:extremelyDif}), the comparison Physiotherapists-Surgeons shows lowest similarity (0.640).
	\item For the linguistic descriptor \textit{Moderately difficult} (see also Fig. \ref{fig:ModeratelyDif}), the comparison Physiotherapists-Surgeons shows the lowest similarity (0.614).
	\item For the linguistic descriptor \textit{A little bit difficult} (see also Fig. \ref{fig:aLittleBitDif}), comparisons between Physiotherapists-Surgeons and Physiotherapists-Patients show low similarity (0.622 and 0.683 respectively).
	\item For the linguistic descriptor \textit{Not at all difficult} (see also Fig. \ref{fig:notAtAllDif}), the comparison Physiotherapists-Surgeons shows the lowest similarity (0.512) amongst all the word models.
\end{itemize}

\begin{figure}[htbp]
	\centering
	\includegraphics[scale=0.62]{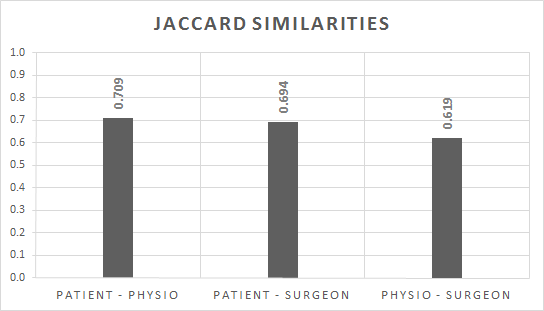}
	\caption{Averages of FS similarities between different groups of people.}
	\label{fig:similarities}
\end{figure}

Table \ref{tab:AveragedSimilarities} and Figure \ref{fig:similarities} show the average of the similarities over the 5 words, thus providing an indication of how similar the interpretation of the TESS item vocabulary is by the key stakeholder groups: Patients, Physiotherapists and Surgeons. While the sample used for this paper is too small to draw any conclusions, it is interesting to note that the understanding is best (most similar) between patients and physiotherapists. While the similarity in interpretation is lowest between physiotherapists and surgeons. Note that as here we are interested in comparing real-world stakeholder groups, the similarity to the generated group of medical professionals is omitted.
% as whole across the different stakeholder groups. Note that, as a result of an increment in agreement between Physiotherapists and Surgeons, the similarity average of Physio-Surgeon group is higher than the averages of both groups. It can also be noted that, in general, the Surgeon group has the lowest agreement with regard to the rest of the groups while the Physiotherapist group is the one with most disagreement. Again, these are not definitive conclusions due to the data sample size however, it is interesting noting the level of disagreement with regard to linguistic descriptors in a relatively small medical professional group working in the same venue.

\begin{table}[htpb]
	\centering
	\caption{Average of FS Similarities across all words}
	\begin{tabular}{r|rrrr}
		\toprule	
	& \textbf{Patient} & \textbf{Physio} & \textbf{Surgeon} & \textbf{Physio-Surgeon} \\
	\midrule
	\textbf{Patient} & \multicolumn{1}{c}{-} & \multicolumn{1}{c}{0.709} & \multicolumn{1}{c}{0.694} & \multicolumn{1}{c}{0.751} \\
	\textbf{Physio} & \multicolumn{1}{c}{0.709} & \multicolumn{1}{c}{-} & \multicolumn{1}{c}{0.619} & \multicolumn{1}{c}{0.773} \\
	\textbf{Surgeon} & \multicolumn{1}{c}{0.694} & \multicolumn{1}{c}{0.619} & \multicolumn{1}{c}{-} & \multicolumn{1}{c}{0.800} \\
	%\textbf{Physio-Surgeon} & \multicolumn{1}{c}{0.751} & \multicolumn{1}{c}{0.773} & \multicolumn{1}{c}{0.800} & \multicolumn{1}{c}{-} \\
		\bottomrule
	\end{tabular}%
	\label{tab:AveragedSimilarities}%
\end{table}%

Table \ref{tab:centroids} shows the centroids of the FSs depicted in Fig. \ref{fig:wordsConceptsFS}, while Table \ref{tab:heights} and Table \ref{tab:supports} show their heights and the size of their support. The centroids, heights and support provide a basic numeric description of the FSs as a means of additional comparison allowing us to deduce the following:
\begin{itemize}
\item For the first linguistic descriptors \textit{Impossible to do} and \textit{Extremely difficult} it can be noted that Patients' and Physiotherapists' centroids are the closest among the groups whereas the Surgeons' centroid is higher considerably.
\item For the linguistic descriptor \textit{Moderately Difficult}, all of the centroids are $ \approx 4.3 $ which can be interpreted as a generalized agreement in the perception of the descriptor \textit{Moderately Difficult}. Interestingly, it can be noted that the FSs are slightly ``balanced'' to the left side of the scale. This can indicate that the descriptor \textit{Moderately Difficult} has not necessarily a neutral meaning as expected.
%\item For the linguistic descriptor \textit{A little bit difficult}, the Patients' and Physiotherapists' centroids are the closest among all models. Therefore, this difference might indicate that when Surgeons have interpreted such descriptor they have understood that the difficulty is slightly higher than the one perceived by both the Patients' and Physiotherapists' perceptions. However, given the relative high uncertainty in these FSs, low agreement overall and the reduced number of data samples we are unable to conclude until performing analysis with more information.
\item For the linguistic descriptor \textit{Not at all difficult}, the centroid associated to the Surgeons' model is considerably lower than the rest. This can be interpreted (along with the width of the FS) to a different perception of the linguistic descriptor for the Surgeons in which it can be used in more varied situations.
\item The support and heights of the FS models related to the \textit{Impossible to do} and \textit{Not at all difficult} descriptors are the narrowest and highest amongst Patients and Physiotherapist, indicating such words are the less ambiguous for those groups.
\item The size of support of the descriptor \textit{Extremely Difficult} was almost identical for all groups of participants indicating that, despite the lack of a high agreement level in general (indicated by the height), the perception of such word can cover the same range of scenarios with all groups.
\item For the case of the linguistic descriptor \textit{A little bit difficult}, it can be seen that in general, all models have a low height (in comparison to other word models) and conversely, the supports are the widest overall covering the major part of the scale. These observations can suggest that the linguistic descriptor \textit{A little bit difficult} does not describe in a satisfactory manner the intended level of function among our population of participants.
\end{itemize}

\begin{table}[htbp]
	\centering
	\caption{Centroids of FS word models from different groups of people}
	\begin{tabular}{r|rrrr}
		\toprule
		& \textbf{Patient} & \textbf{Physio} & \textbf{Surgeon} & \textbf{PS} \\
		\midrule
		\textbf{Impossible to do} & \multicolumn{1}{c}{0.686} & \multicolumn{1}{c}{0.727} & \multicolumn{1}{c}{0.988} & \multicolumn{1}{c}{0.878} \\
		\textbf{Extremely Difficult} & \multicolumn{1}{c}{1.711} & \multicolumn{1}{c}{1.767} & \multicolumn{1}{c}{2.085} & \multicolumn{1}{c}{1.954} \\
		\textbf{Moderately Difficult} & \multicolumn{1}{c}{4.356} & \multicolumn{1}{c}{4.312} & \multicolumn{1}{c}{4.289} & \multicolumn{1}{c}{4.290} \\
		\textbf{A little bit difficult} & \multicolumn{1}{c}{6.433} & \multicolumn{1}{c}{6.462} & \multicolumn{1}{c}{6.126} & \multicolumn{1}{c}{6.188} \\
		\textbf{Not at all difficult} & \multicolumn{1}{c}{9.051} & \multicolumn{1}{c}{9.279} & \multicolumn{1}{c}{8.555} & \multicolumn{1}{c}{8.814} \\
		\midrule
		\textbf{Overall (Average):} & \multicolumn{1}{c}{4.448} & \multicolumn{1}{c}{4.509} & \multicolumn{1}{c}{4.408} & \multicolumn{1}{c}{4.425} \\
		\bottomrule
	\end{tabular}%
	\label{tab:centroids}%
\end{table}%

\begin{table}[htbp]
	\centering
	\caption{Heights of FS word models from different groups of people}
	\begin{tabular}{r|rrrr}
		\toprule
		& \textbf{Patient} & \textbf{Physio} & \textbf{Surgeon} & \textbf{PS} \\
		\midrule
		\textbf{Impossible to do} & 1.000     & 1.000     & 0.917 & 0.875 \\
		\textbf{Extremely Difficult} & 0.917 & 0.769 & 0.917 & 0.792 \\
		\textbf{Moderately Difficult} & 0.750  & 0.923 & 1.000     & 0.958 \\
		\textbf{A little bit difficult} & 0.667 & 0.615 & 0.583 & 0.542 \\
		\textbf{Not at all difficult} & 1.000     & 1.000     & 0.833 & 0.875 \\
		\bottomrule
	\end{tabular}%
	\label{tab:heights}%
\end{table}%

\begin{table}[htbp]
	\centering
	\caption{Size of support of FSs related to word models from different groups of people}
	\begin{tabular}{r|rrrr}
		\toprule
		& \textbf{Patient} & \textbf{Physio} & \textbf{Surgeon} & \textbf{PS} \\
		\midrule
		\textbf{Impossible to do} & 2.1   & 2.4   & 3.2   & 3.2 \\
		\textbf{Extremely Difficult} & 4.3   & 4.2   & 4.2   & 4.2 \\
		\textbf{Moderately Difficult} & 6.3   & 5.5   & 4.8   & 6.2 \\
		\textbf{A little bit difficult} & 7.3   & 7.6   & 8     & 7.7 \\
		\textbf{Not at all difficult} & 2.5   & 2.4   & 4.5   & 4.5 \\
		\bottomrule
	\end{tabular}%
	\label{tab:supports}%
\end{table}%

\section{Discussion}
The IAA is designed to avoid assumptions about the data or FS shape \cite{Wagner2014}, instead, it models the agreement/disagreement (in \textit{y}) and the uncertainty associated (in \textit{x}) from multiple intervals capturing a linguistic term. In this paper, we have used it to create FSs of the standardised vocabulary used in the TESS by three key stakeholder groups.

We began this study with the question if there are differences in the perception of the linguistic descriptors in the TESS that may influence the medical assessment and therapy outcomes. This analysis, despite being at a preliminary stage with limited data, suggests that each of the terms is not necessarily equidistant as assumed in the items of a Likert scale. Also, the similarity values indicate that for some stakeholders, the interpretations of linguistic descriptors vary considerably. %Moreover, numerical-based comparisons have shown that the perceptions of Surgeons about the linguistic descriptors used in TESS are the most different with regard to the perceptions of Physiotherapists and Patients.

As expected, all FSs present the same order i.e., when comparing the ``order of appearance'' in each group of participants the first FS is \textit{Impossible to do}, the second FS is \textit{Extremely difficult} and so on. Therefore, the items (linguistic descriptors) do follow an ascending order in a ranking sense. However, by looking at the centroids of the FSs we can detect that they are not equidistant since their differences would be expected to be (at least) approximately equal. For example, the minimum and maximum differences between linguistic models created from Physiotherapists' responses are 1.041 (\textit{Impossible to do} and \textit{Extremely difficult}) and 2.818 (\textit{A little bit difficult} and \textit{Not at all difficult}). In addition, comparisons between the sizes of support help to show the differences in the spread of each word's interpretation model on the scale, where the size of the descriptor \textit{A little bit difficult} is considerably wider than the rest of the words (across all groups).

In general, the linguistic descriptors found at the boundaries (i.e., \textit{Impossible to do} and \textit{Not at all difficult}) and in the middle of the scale (\textit{Moderately difficult}) are more defined and have higher agreement whereas the remaining descriptors have more uncertain distributions. Here, \textit{A little bit difficult} is the model reflecting the lowest level of agreement in all cases and the highest uncertainty, this, if confirmed with a more substantial dataset, can indicate that these terms lead to potential miscommunication between stakeholders of the TESS.%Also, their associated centroids are quite approximated meaning that, in all participant groups it is agreed that the descriptor \textit{A little bit difficult} is the most disaccord and uncertain for these participants.

\section{Conclusions and Future Work}
In this study, we have modelled the perceptions of the linguistic descriptors used as the standard vocabulary in the TESS questionnaire using the Interval Agreement Approach. We used an interval-valued scale to capture the inter-participant uncertainty about the extent of the given descriptors with 4 groups, namely: Patients, Surgeons, Physiotherapist and Surgeons-Physio, i.e., Health Professionals.

We performed comparisons between the Fuzzy Sets, showing that some words are clearer and more unanimously understood than others. In particular, we found very little similarity between Physiotherapists and Surgeons for the descriptors \textit{Extremely difficult}, \textit{Moderately difficult}, \textit{A little bit difficult} and \textit{Not at all difficult}. We also analysed the centroids of the models as a means of numeric description indicating that the items are not equidistant (as assumed for the Likert scale). In particular, for the specific case of the term \textit{A little bit difficult} it is interesting how it stands out as the least well defined, with generally low agreement and at times, a quasi bi-modal model and interpretation.

% something wrong with ALBD, focus for the future, is the one that stands out in being the least well defined
%want to continue with more data
%is bi-modal, it might have two different interpretations
While this study is at a preliminary stage with a small sample not large enough to support statistically relevant conclusions, the paper has shown that interval-valued data capture and subsequent modelling of the intervals using the IAA provides a promising tool for analysing standardised vocabulary -for example, in medical-patient communication. Such analysis can be used to identify potential variations in meaning/perceptions of key words in medical treatment/diagnosis and thus support improved expert-patient communication which in turn, may lead to improve medical treatment.%of the need for modifying (a non-fixed number of words) can be assessed by considering the stakeholders' perceptions.
%assess  and find descriptive and standardised vocabulary in several contexts (e.g., patient consent, risk communication, quality of life assessment)
% * <bulerias.snow.toy@gmail.com> 2016-03-23T13:29:01.374Z:
%
% >  Such analysis, can be used to assess potential variations in meaning/perceptions of key words and find descriptive and standardised vocabulary in several contexts (e.g., patient consent, risk communication, quality of life assessment) which in turn, may lead to improve medical treatment.
%
% Added as response to reviewers 2 and 3.
%
% ^.

In the future, we aim to collect a larger sample of the proposed data to validate the initial finding in this paper. More fundamentally, we are seeking to develop the approach used in this paper to experimentally assess the equidistant spacing of terms on ordinal scales while exploring the additional information provided by interval-valued scales. In relation to the latter, we are in particular looking to transform questionnaires such as the TESS to an interval-valued scale and to analyse the outcome and its potential for real-world application. Finally, we are currently developing further tools for the (statistical) comparison of data-driven fuzzy sets such as generated by the IAA.
%In the future, we aim to generate qualitative descriptions of the patients' physical functions similarly from patients, surgeons and physiotherapists using more data and as such, develop a more representative analysis. Similarly, we plan to gather data from patients related to their difficulty perceptions to perform daily activities based on the TESS questionnaire. Such information will enable us to develop better assessments of the patients' functions by allowing a flexible capture of participants' perceptions.
%\appendix[Questionnaire structure]

\section*{Acknowledgement}
The authors would like to thank all those medical professionals and patients from Nottingham University Hospital (City Campus) who kindly took part in the survey used in this study.
\ifCLASSOPTIONcaptionsoff
  \newpage
\fi


\bibliography{library}

%\begin{biography}[{\includegraphics[width=1in,height=1.25in,clip,keepaspectratio]{mshell}}]{Michael Shell}


%\vfill


\end{document}